%% file: main.tex
\documentclass[10pt,twocolumn]{article}
\usepackage{lineno}
\usepackage{graphicx}
\usepackage{amsmath,amssymb}
\usepackage{times}
\usepackage{epsfig}
\usepackage{booktabs}
\usepackage{url}
\usepackage{caption}
\usepackage{subcaption}
\usepackage{lipsum} 
\usepackage[margin=0.9in,top=0.7in,columnsep=20pt]{geometry}
\usepackage{xcolor}  
\usepackage{hyperref}  
\usepackage{bm}        
\usepackage{amsfonts}  
\newcommand{\G}{\Gamma}
\newcommand{\bs}[1]{\bm{#1}} 
\newcommand{\front}{\text{front}}   
\newcommand{\back}{\text{back}}     
\newcommand{\gt}{\text{gt}}     

\hypersetup{
    colorlinks=true,    
    linkcolor=blue,     
    urlcolor=blue       
}

\title{\textbf{NI-Tex: Non-isometric Image-based Garment Texture Generation}}

\author{
Hui Shan$^{1,2,3}$,
Ming Li$^{1,2,3}$,
Haitao Yang$^{4}$,
Kai Zheng$^{3}$,
Sizhe Zheng$^{2,3}$,\\
Yanwei Fu$^{2,5}$,
Xiangru Huang$^{3}$$^\dagger$\\[3pt]
$^1$Zhejiang University \quad
$^2$Shanghai Innovation Institute \quad
$^3$Westlake University \\
$^4$University of Texas at Austin \quad
$^5$Fudan University
}

\date{} 
\renewenvironment{abstract}
 {
  \begin{center}
    \Large\bfseries Abstract  
  \end{center}
  \begin{itshape}
 }
 {
  \end{itshape}
 }

\begin{document}
\twocolumn[{
\begin{center}
\maketitle
\end{center}

\begin{center}
\includegraphics[width=\linewidth]{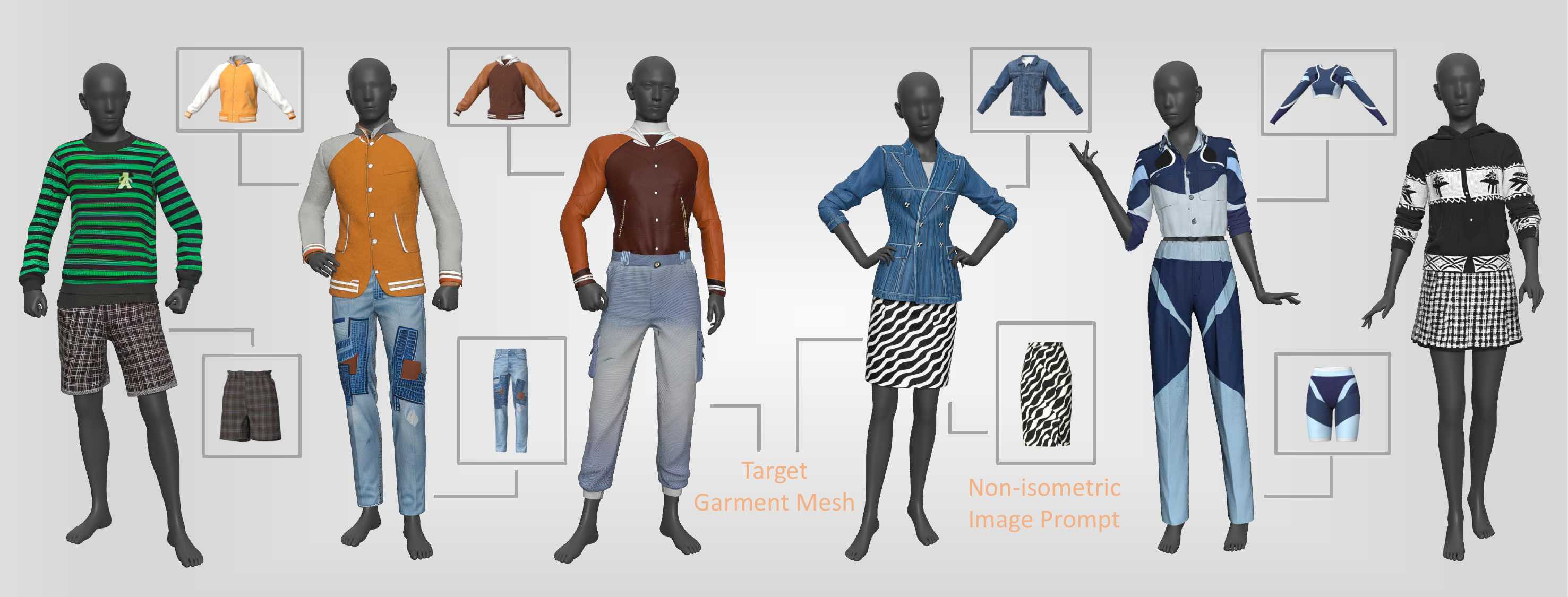}
\captionof{figure}{NI-Tex takes an image prompt and a target garment geometry as input, and generates high-quality PBR textures that faithfully transfer the textures and patterns from the input image to the target garment. Unlike other existing methods, the generation quality of NI-Tex does not degrade for challenging image-garment pairs with strong topological and geometric differences, enabling superior flexibility.}
\label{fig:teaser}
\vspace{0.3cm}
\end{center}
}]

\begingroup
\renewcommand{\thefootnote}{\textdagger} 
\footnotetext{\hangindent=1.8em \hangafter=1 Corresponding author (huangxiangru@westlake.edu.cn).}
\endgroup

\input{section/0.abstract}
\input{section/1.Intro}
\input{section/2.RelatedWork}
\input{section/3.Preliminary}

\input{section/4.Method}

\input{section/5.Experiments}

\input{section/6.Conclusions}

{
\small
\bibliographystyle{unsrt}
\bibliography{main}
}

\appendix
\input{section/Appendix}

\end{document}

%% file: section/0.abstract.tex
\begin{abstract}
Existing industrial 3D garment meshes already cover most real-world clothing geometries, yet their texture diversity remains limited. To acquire more realistic textures, generative methods are often used to extract Physically-based Rendering (PBR) textures and materials from large collections of wild images and project them back onto garment meshes. However, most image-conditioned texture generation approaches require strict topological consistency between the input image and the input 3D mesh, or rely on accurate mesh deformation to match to the image poses, which significantly constrains the texture generation quality and flexibility.

To address the challenging problem of non-isometric image-based garment texture generation, we construct 3D Garment Videos, a physically simulated, garment-centric dataset that provides consistent geometry and material supervision across diverse deformations, enabling robust cross-pose texture learning. We further employ Nano Banana for high-quality non-isometric image editing, achieving reliable cross-topology texture generation between non-isometric image-geometry pairs. Finally, we propose an iterative baking method via uncertainty-guided view selection and reweighting that fuses multi-view predictions into seamless, production-ready PBR textures. Through extensive experiments, we demonstrate that our feedforward dual-branch architecture generates versatile and spatially aligned PBR materials suitable for industry-level 3D garment design. The code will be released on \url{https://github.com/SII-Hui/NI-Tex}.

\end{abstract}

%% file: section/1.Intro.tex
\section{Introduction}
\label{sec:intro}

Acquiring high-quality 3D garment assets is important for a number of applications such as virtual reality, human avatars and physical simulation. While existing garment datasets already cover a rich distribution of geometries, a key question is how to augment existing garment geometries with more diverse and photorealistic Physically-based Rendering (PBR) textures. A common strategy to acquire more realistic textures is to employ texture generation methods that synthesize desired PBR textures conditioning on a target geometry and an user-provided image prompt, which offers precise control over the generated textures.

\begin{figure}[h]
    \centering
    \includegraphics[width=\linewidth]{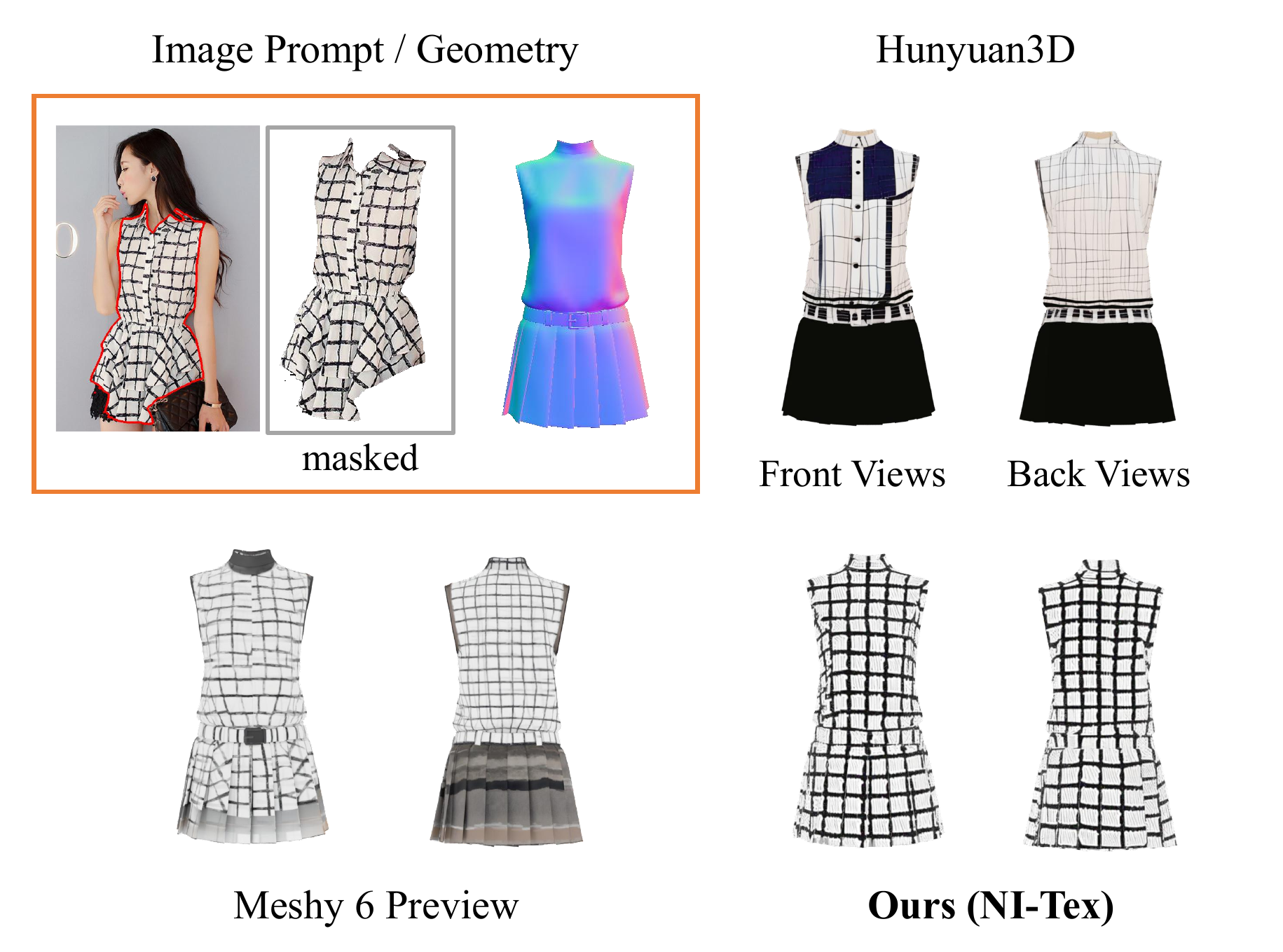}
    \caption{Texture generation becomes unreliable when the image prompt and the given mesh exhibit topology inconsistency in Hunyuan3D~\cite{hunyuan3d2025hunyuan3d} and Meshy 6 Preview.}
    \label{fig:DemoforIntro}
\end{figure}

Current texture generation models are typically trained on large-scale 3D datasets such as Objaverse~\cite{deitke2023objaverse} and TexVerse~\cite{zhang2025texverse}, which provide massive collections of 3D assets with PBR textures. These models usually use multi-view attention mechanisms to learn implicit image-geometry correspondences to faithfully transfer the visual appearance and material properties from the image to the target geometry. However, such correspondence learning relies heavily on geometric and topological similarity between the input image and the target geometry. When there exists a significant topological or geometric discrepancy, especially in cases of non-isometric deformation or garment topology variation, the generation quality deteriorates drastically, leading to distorted or inconsistent textures like Figure~\ref{fig:DemoforIntro}.

In this work, we address this critical challenge by proposing \textbf{NI-Tex}, a training framework that promotes non-isometric image-based garment texture generation, which focuses on robust and controllable texture generation under topological and geometric image-garment inconsistencies. NI-Tex builds upon a texture generation backbone pretrained on Objaverse and TexVerse, and we further fine-tune it using datasets with rich garment deformation data such as BEDLAM~\cite{black2023bedlam} to improve the generalization for garment deformations. 

To accommodate for topological image-garment differences, we introduce a novel data augmentation strategy that leverages image editing tools (e.g., Nano Banana) to modify the topology and geometry of the image prompt. We propose a set of carefully designed strategies to regulate this process for multiple purposes: 1) ensure diverse topology modifications; 2) reduce the sim-to-real gap when applying image editing models; and 3) preserve the texture consistency before and after modification. This process results in a new garment texture dataset with a special enhancement for non-isometric image-garment pairs. Extensive qualitative and quantitative experiments suggest that NI-Tex significantly outperforms existing baselines with superior robustness and generation quality under challenging topological and geometric differences.

Beyond dataset and model design, we further propose an iterative baking algorithm to improve the stability of multi-view texture integration. Specifically, our train an Uncertainty Quantification (UQ) model to predict the per-pixel uncertainty on the generated texture maps. This uncertainty representation is capable of capturing a wide range of baking artifacts, such as holes and blurryness. To acquire the training data for the UQ model, we employ the trained texture generation models and simulate the baking error on another set of edited image prompts. The uncertainty annotation can be easily collected by compared the generated textures with the ground-truth texture maps. 

To fix the baking artifacts, we iteratively select new views that maximize the average per-pixel uncertainty, and apply simple reweighting strategy to robustly merge multi-view generated textures into integrated mesh textures.

To summarize, our contributions are:
\begin{itemize}
    \item We are the first to adopt a feedforward architecture to effectively tackle the challenging task of non-isometric image-based texture generation. A novel image editing based framework for addressing non-isometric image-garment inconsistencies, effectively leveraging powerful image editing tools to enhance controllability and robustness of image-based garment texture generation.
    \item A new dataset designed to improve the generation quality for non-isometric image garment pairs, which will be publicly released to facilitate future research.
    \item An uncertainty-aware iterative baking algorithm that improves the stability and quality of multi-view texture integration.
\end{itemize}

%% file: section/2.RelatedWork.tex
\section{Related work}
\label{sec:related work}

\subsection{3D Texture Generation}
3D Texture Generation methods study how to add textures to given 3D geometries (represented by mesh) from user-defined text or image prompts. TEXTure~\cite{richardson2023texture} and Text2Tex~\cite{chen2023text2tex} generate images from different viewpoints using diffusion models. These images are then back-projected onto the surface of the geometry. To further improve the alignment between the generated images and the given geometry, Easi-Tex~\cite{perla2024easi} uses edge maps to guide the diffusion process. However, images from different viewpoints often cannot fully cover the entire geometry. Thus, some faces of the mesh remain untextured. This problem is more evident for objects with complex structures. Paint3D~\cite{zeng2024paint3d} additionally introduces an inpainting step on the UV map of the geometry, allowing all faces to be textured. By incorporating global texture information, it further improves the texture consistency across multiple viewpoints. Although the above methods achieve good results in 3D texture generation, they overlook the process of lighting and shadows in the generated images. As a result, the produced textures fail to deliver realistic visual effects under new lighting conditions.

\subsection{PBR Texture Generation}
PBR texture generation methods output the object’s base color (albedo) and its perceptual properties (metallic and roughness) separately. The Score-Distillation-Sampling (SDS) based optimization approach and the data-driven feed-forward approach are two common categories for PBR texture generation. Fantasia3D~\cite{chen2023fantasia3d}, Matlaber~\cite{xu2023matlaber}, and Paint-it~\cite{youwang2024paint} leverage the priors of pretrained 2D diffusion models and use SDS to distill the PBR texture attributes of the object. FlashTex~\cite{deng2024flashtex} and DreamMat~\cite{zhang2024dreammat} train light ControlNet to achieve light-aware diffusion models by augmenting the training dataset with simulated reflection effect under various lighting conditions. However, SDS-based optimization is time-consuming and sometimes produces less realistic visual effects. Another line of research focuses on feed-forward training enabled by existing open-source 3D datasets~\cite{deitke2023objaverse,deitke2023objaverse-xl,zhang2025texverse}. MaterialAnything~\cite{huang2025material} and Hunyuan3D 2.1~\cite{hunyuan3d2025hunyuan3d} render PBR-annotated objects from the datasets (mainly from Objaverse) under different lighting conditions and viewpoints. They use albedo, metallic, and roughness maps for supervision in training, which makes the model material-aware. The feed-forward approach effectively reduces inference time, enabling efficient and large-scale PBR texture generation. Hunyuan3D 2.1 applies rigid operations such as scaling, translation, and rotation during rendering to augment training data for better generalization. However, the model still struggles to generate textures between image-geometry pairs with non-isometric differences such as topology changes. 

\subsection{Image-based Virtual Try-Off and Image-Prompt 3D Garment Texture Generation} TryOffDiff~\cite{velioglu2024tryoffdiff} is the first to introduce Virtual Try-Off (VTOFF), which extracts garment texture from given photos of clothed individuals and generates standardized 2D garment images. OMFA~\cite{liu2025one} and Voost~\cite{lee2025voost} build a unified framework that combines Virtual Try-On (VTON) and VTOFF techniques to achieve better results. However, they are limited to generating 2D images and cannot bake the extracted garment textures onto a given 3D geometry. In the field of image-based 3D garment generation, methods such as Pix2Surf~\cite{mir2020learning}, Cloth2Tex~\cite{gao2024cloth2tex}, and Garment3DGen~\cite{sarafianos2025garment3dgen} typically require a 2D garment image with canonical pose and then align a template mesh to the garment image through mesh deformation, which suffers from compound errors from the mesh alignment procedure and has limited applicability.

%% file: section/3.Preliminary.tex
\begin{figure*}[t]
    \centering
    \includegraphics[width=0.85\linewidth]{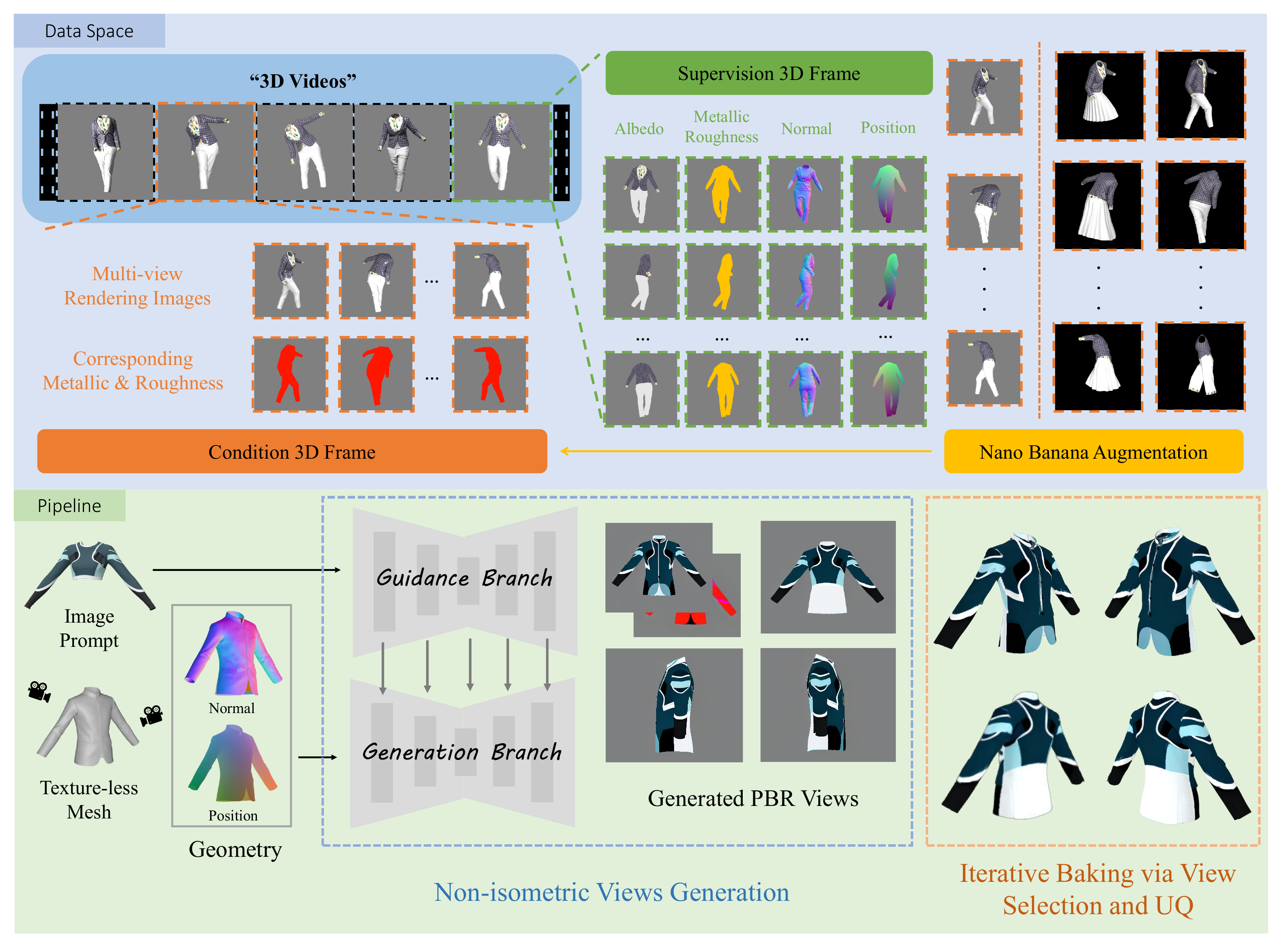}
    \caption{\textbf{Overview of NI-Tex.} In the data space (top), we construct our non-isometric training dataset from 3D Garment Videos by randomly selecting two frames, one as the condition 3D frame and the other as the supervision 3D frame, to enhance the model’s generalization across human poses, geometric deformations, and lighting variations. To further improve robustness to different garment topology, we apply Nano Banana for image editing on randomly rendered views of the condition 3D frame. In the pipeline (bottom), we render the input texture-less mesh to obtain normal and position maps as geometric constraints and employ a dual-branch architecture to achieve non-isometric PBR view generation. Finally, view selection and Uncertainty Quantification (UQ) are used to iteratively bake the results across multiple viewpoints.}
    \label{fig:overview}
\end{figure*}

\section{Preliminary}
\label{sec:preliminary}

NI-Tex is designed for image-based garment texture generation with a special enhancement for non-isometric image-garment pairs. In this section, we first formally define the task of image-based garment texture generation (section~\ref{subsec:problem:setup}), then briefly introduce the backbone network architecture that we adopted from Hunyuan3D (section~\ref{subsec:network}).

\subsection{Problem Setup}
\label{subsec:problem:setup}
Image-based garment texture generation takes an input RGB image $I \in \mathbb{R}^{H\times W \times 3}$ and an input geometry represented by a mesh, and output multiple generated texture maps (albedo, roughness, metallic). Each texture map is defined as an image $a\in \mathbb{R}^{N \times N \times C}$ in the UV space of the input geometry. Albedo, roughness, metallic texture map contains $C=3,1,1$ channels, respectively. 

\subsection{Network Architecture}
\label{subsec:network}
To preserve the texture information from the input image prompt during training, we adopt a feedforward dual-branch architecture composed of a guidance branch and a generation branch like Hunyuan3D. The guidance branch extracts hierarchical features from the input image and feeds them into the corresponding layers of the training branch as reference information. The training branch takes multi-view normal and position maps as input, while leveraging the reference features to perform multi-channel inference of albedo and MR (metallic and roughness) attributes.

\paragraph{Multi-Channel Aligned Attention (MCAA)} 

 We adopt the MCAA module from MaterialMVP \cite{he2025materialmvp} to connect the two branches. 
As defined in Equation (\ref{eq:MCAA}), we extract the reference texture features from the guidance branch and inject them into the albedo channel of the training branch:
\begin{equation}
\text{Attn}_{albedo} = \text{Softmax}\left( \frac{Q_{albedo} K_{ref}^{T}}{\sqrt{d}} \right) \cdot V_{ref}
\label{eq:MCAA}
\end{equation}

To achieve spatial and geometric alignment between the MR and albedo attributes, we inject the albedo attention into the MR latent representation:
\begin{equation}
z_{MR}^{new} = z_{MR} + \text{Attn}_{albedo},
\label{eq:MR-Aligned}
\end{equation}

%% file: section/4.Method.tex
\section{Non-isometric Garment Texture Generation}
\label{sec:method}

NI-Tex proposes multiple novel strategies to enhance the generation quality for non-isometric image-garment pairs. First, to prepare training datasets for non-isometric image-garment pairs, we propose to use simulation data and image editing techniques to mitigate the generalization gap in terms of human pose, garment topology, lighting and geometric deformation (section~\ref{subsec:dataset:preparation}). To compensate for severe geometric and topological differences in test time, which can cause artifacts such as holes and view-inconsistent textures, we also propose an iterative baking procedure based on uncertainty quantification and view-selection (section~\ref{subsec:baking}). Finally, in section~\ref{subsec:implementation:details}, we elaborate the training architecture of NI-Tex.


\subsection{Dataset Preparation}
\label{subsec:dataset:preparation}
In this section, we introduce how we construct our training dataset for enhancing non-isometric image-garment pairs. 

\noindent\textbf{Cross-pose Augmentation.} 
We construct 3D Garment Videos based on the \textbf{BEDLAM}\cite{black2023bedlam} dataset. 
BEDLAM contains nearly 1,700 distinct garment albedo textures, each associated with hundreds of motion sequences, and each sequence consists of hundreds of frames. Since the original dataset does not provide roughness or metallic attributes, we augment the selected motion sequences by assigning PBR material properties to each frame. Specifically, we set roughness and metallic values as: 
\begin{equation}
\text{roughness} \sim \mathcal{U}(0,1),~\quad \text{metallic} = 0
\end{equation}

For each physically-simulated motion sequence in BEDLAM, we extract per-frame garment geometry and form a sequence $V = \{M_1, M_2, M_3, \dots, M_n\}$. The geometric deformations of each garment $M_i$ are driven by coherent temporal human motion, with a consistent albedo texture map shared across all frames (See Figure~\ref{fig:overview}). We also denote $a$ as the albedo texture map. We observe that fabric specular appearance is mainly determined by roughness, while metallic maps have little visual effect. Additionally, a single garment usually has uniform material properties, so its PBR attributes should remain consistent.

\begin{figure*}[t]
    \centering
    \includegraphics[width=0.97\linewidth]{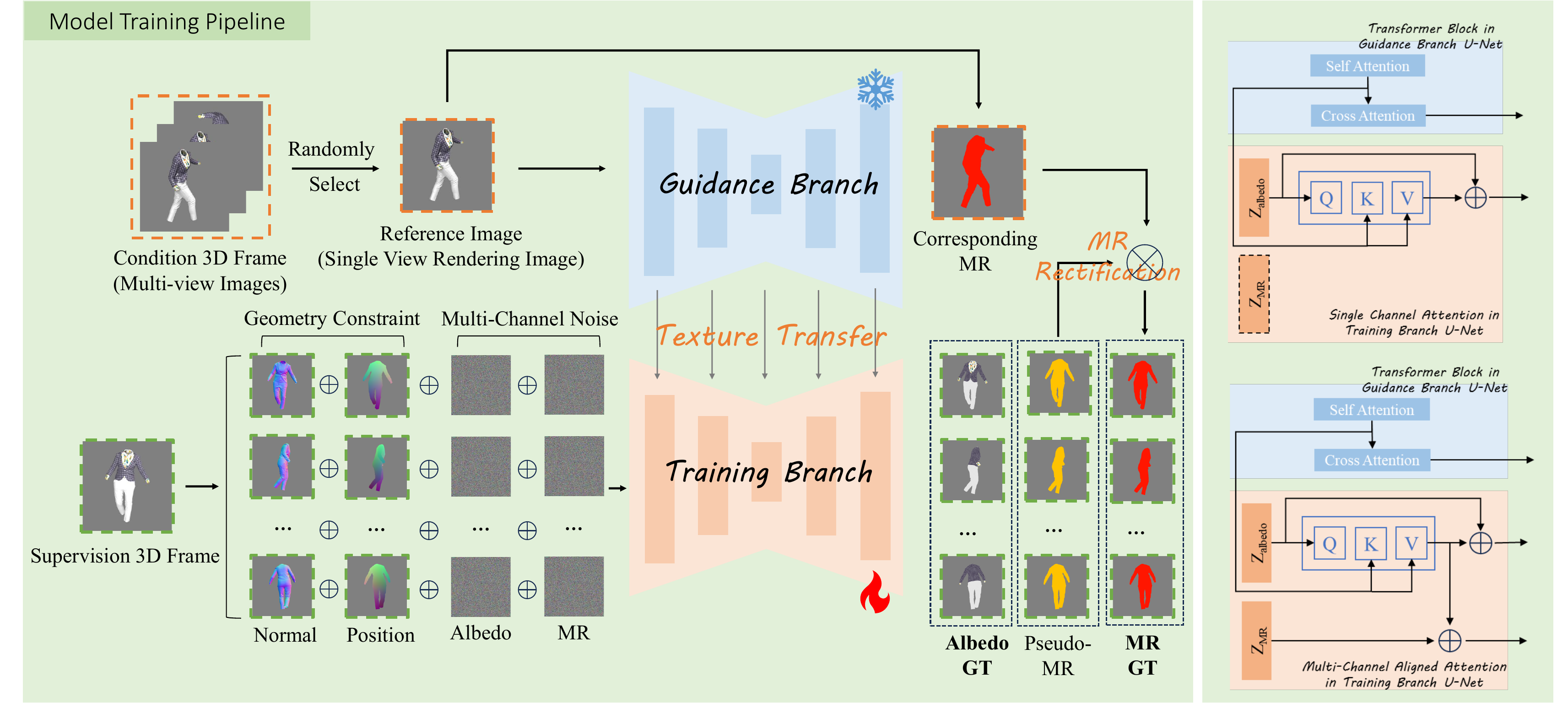}
    \caption{\textbf{(left)} In the model training pipeline, we use an illuminated image from the condition 3D frame as the reference image, which is encoded by the guidance branch to perform texture transfer to the training branch. The supervision 3D frame provides multi-view normal and position images as geometric constraints, and its PBR (albedo and MR) maps for supervision. During supervision, MR Rectification is applied to the MR maps of the supervision frame to obtain refined ground truth.
    \textbf{(right)} The architecture of our switchable multi-channel U-Net. Since images generated by Nano Banana cannot maintain consistent MR properties, we apply single-channel attention for albedo-only supervision (right top), and multi-channel aligned attention for joint albedo–MR supervision (right bottom).}
    \label{fig:modelArchitecture}
\end{figure*}

During training, we randomly sample two frames from each motion sequence (Figure~\ref{fig:modelArchitecture}, left). One frame is used as the \textbf{condition 3D frame} and the other as the \textbf{supervision 3D frame}. 
For the condition frame, we select one illuminated view as the input image prompt and use its normal and position maps from ten viewpoints as geometric constraints. 
For the supervision frame, we use its PBR texture attributes from ten viewpoints as supervision. 
This cross-frame supervision enables the model to learn texture consistency and variations across poses, resulting in more stable texture generation. 
Moreover, since any two frames in a 3D garment video can form a valid pair, the dataset expands combinatorially, scaling hundreds of thousands of frames into tens of billions of training samples.



\noindent\textbf{Cross-topology Augmentation.} To enable texture generation across diverse garment topologies (e.g., generating textures from skirts or shorts images onto long-pants meshes), we randomly sample rendered views from 3D Garment Videos and apply \textbf{Nano Banana} to edit garment topology while preserving the original textures. 
We use illuminated renderings instead of albedo images, as they yield more realistic results and reduce the domain gap during inference. 
The edited images replace the original condition images, while supervision remains from the original supervision 3D frames. 
This image–garment training effectively distills texture identity consistency from Nano Banana. 
To avoid incorrect distillation, we preserve the semantic integrity of garment textures following three key principles (Figure~\ref{fig:NBDemo}):

\begin{figure}[h]
    \centering
    \includegraphics[width=0.9\linewidth]{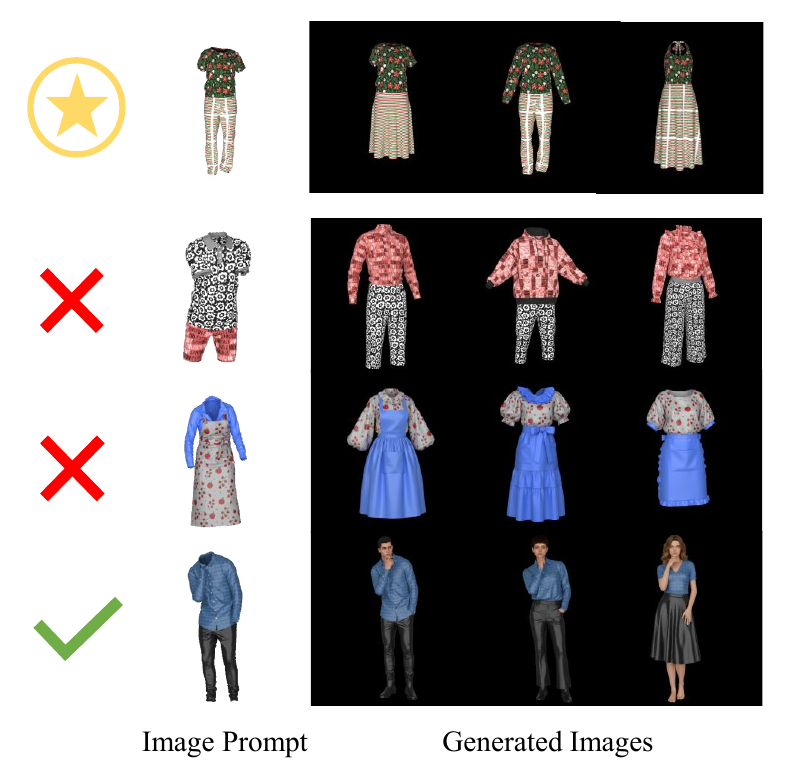}
    \caption{We use Nano Banana to edit garment topology while preserving texture consistency (first row). We ensure category-wise and inner–outer texture consistency to avoid texture swaps or layering confusion (second and third rows). Additional human body generations (fourth row) are acceptable, as image–garment training improves NI-Tex’s understanding of garment textures.}
    \label{fig:NBDemo}
\end{figure}

\begin{enumerate}
    \item \textbf{Category-wise consistency.} When editing full-body garments, textures of upper and lower parts (e.g., tops and skirts or pants) should not drift or swap.
    \item \textbf{Inner–outer consistency.} For layered outfits, outer garment textures must remain distinct from inner ones.
    \item \textbf{Allowance for auxiliary human parts.} We allow Nano Banana to occasionally generate extra human regions, which encourages the model to focus on garment materials and ignore irrelevant areas in wild images.
\end{enumerate}

\begin{figure}[h]
    \centering
    \includegraphics[width=\linewidth]{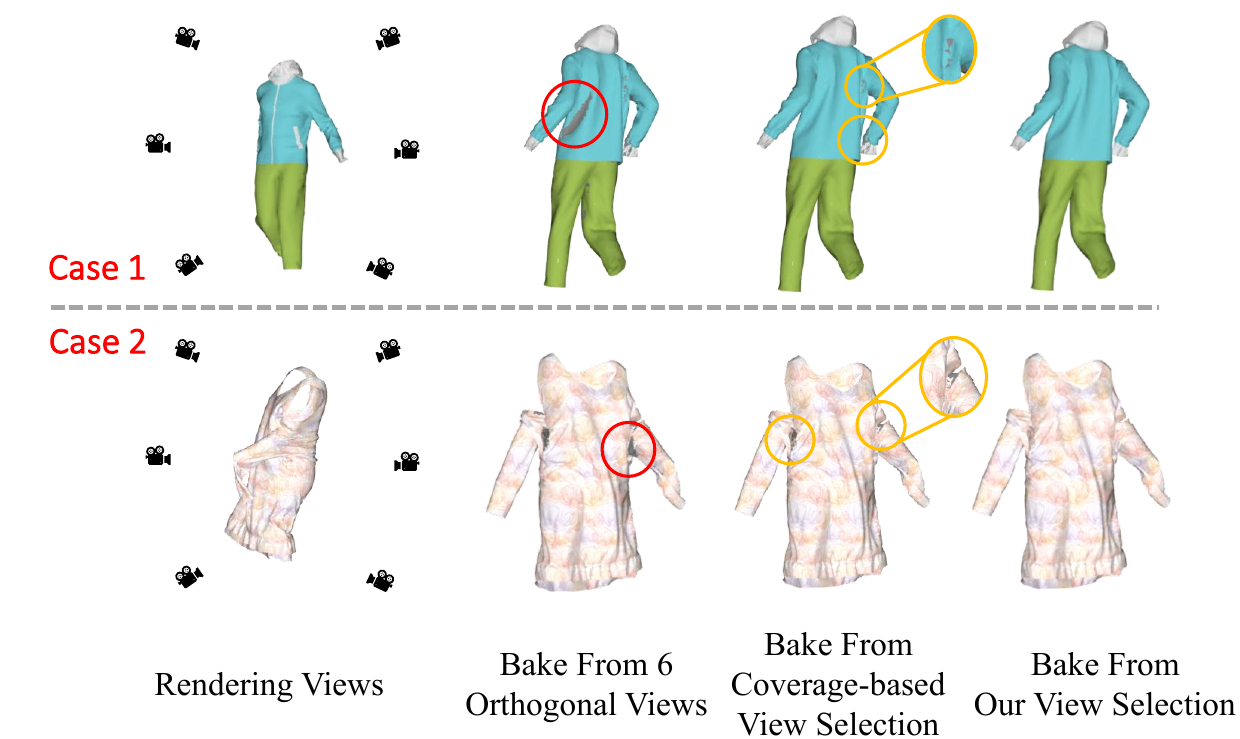}
    \caption{We render each garment mesh from six viewpoints and bake the results to check coverage. Self-occlusion under orthogonal views leaves many regions missing. Coverage-based view selection improves coverage but still misses small areas, while ours achieves full mesh coverage.}
    \label{fig:DemoForViewSelect}
\end{figure}

\subsection{Iterative Baking via View Selection and Uncertainty Quantification}
\label{subsec:baking}
While the generative model trained from the augmented dataset (section~\ref{subsec:dataset:preparation}) produces reasonable multi-view texture maps. The final step of 3D garment texture generation is to merge the multi-view texture maps into the view-independent texture maps on the input mesh. In this paper, we denote this procedure as \emph{baking}. The baking procedure typically introduces unacceptable visual artifacts, such as blurryness or holes due to incompleteness or inconsistency in the predicted multi-view texture maps. 

To fix the artifacts, we first train an Uncertainty Quantification (UQ) model that predicts unsatisfactory areas in the multi-view texture maps. We then propose an UQ-based view-selection algorithm to augment more views at test time. The multi-view texture map predictions in all  (original and augmented) views are weighted by the predicted uncertainty to produce robust texture maps on the mesh. The process continues in an iterative way until no more progress can be made. In the following, we elaborate the details of each step.

\noindent\textbf{Uncertainty Quantification.} We formulate the Uncertainty Quantification task as an Image Quality Assessment (IQA) problem following Active View Selector (AVS)~\cite{wang2025active}. Given an single-view texture map $I\in \mathbb{R}^{H\times W\times 3}$, the UQ model adopts an residual net~\cite{he2016deep} architecture and learns to predict per-pixel uncertainty scores $S \in \mathbb{R}^{H \times W}$ on the multi-view texture maps via supervised learning. 

The training data for UQ model is collected via an error simulation procedure: for a ground-truth mesh and its associated texture maps. We render the textured mesh in 10 views (including the front and back views) and use Nano Banana to edit a random view. The image editing procedure is identical to the one defined in section~\ref{subsec:dataset:preparation}. We then apply our trained texture generation model using the edited image as the image prompt, and optimize the latent code in the diffusion procedure to enforce an as-close-as-possible match between the predicted front and back view texture maps and the corresponding ground truth texture maps. This forms an optimization problem:

\begin{equation}
    \min_{\bs{z}} \|\G^{\front}(\bs{z}) - T^{\front}_{\gt} \|^2 + \|\G^{\back}(\bs{z}) - T^{\back}_{\gt} \|^2
\end{equation}
where $\bs{z}$ is the latent code, $\G^{(\cdot)}$ maps the latent code to the predicted front- or back-view texture map, $T^{(\cdot)}_{\gt}$ represent the ground truth texture maps in front or back view.
Once the problem is solved, we collect the predicted texture maps and the ground truth maps in all views, forming 10 pairs of texture maps. Each pair is automatically associated with a per-pixel uncertainty map computed by comparing each pair of texture maps in terms of SSIM. 

\noindent\textbf{View Selection.} The view selection algorithm takes a set of predefined candidate views and compute a uncertainty score of each candidate view by averaging the per-pixel uncertainty. The candidate view with the maximal uncertainty score is selected by the algorithm, and we make another inference using the multi-view texture generation model for all views, including the one we just selected. Compared to coverage-based view-selection methods, our UQ models is capable of capturing intermediate baking errors and artifacts during the baking procedure, which leads to superior baking results (See section~\ref{sec:qualitativeEvaluation}).

\noindent\textbf{Implementation details.} The iterative process stops when we reach a maximum number of views $N_{\textup{view}}$ or the newly selected view has a uncertainty score that is lower than a threshold $\epsilon$. More details are deferred to Appendix.

\subsection{Training Architecture}
\label{subsec:implementation:details}
\paragraph{Switchable Multi-channel U-Net.}
Images generated by Nano Banana often exhibit inconsistent surface reflections, making it unnecessary to optimize their MR attributes during training. To improve training efficiency, we introduce the switchable multi-channel U-Net (see Figure \ref{fig:modelArchitecture}, right), which allows the MR channel to be deactivated when only albedo optimization is required.

\noindent\textbf{Training Loss.}
The model $\epsilon_{\theta}$ generates two diffusion noises $\epsilon_{t}^{MR}$ and $\epsilon_{t}^{Albedo}$ conditioned on the input image $I$ at training timestep $t$. In the multi-channel optimization stage, loss can be designed as:
\begin{equation}
 \mathcal{L}{_\text{1}}=\mathbb{E}_{\epsilon\sim\mathcal{N}(0,1),\,t}\left[\left\|\epsilon-\epsilon_{t}^{MR}\right\|_{2}^{2}+\left\|\epsilon-\epsilon_{t}^{Albedo}\right\|_{2}^{2}\right]
\end{equation}
When the switchable multi-channel U-Net deactivates the MR channel, the optimization is performed solely on the albedo channel:
\begin{equation}
 \mathcal{L}{_\text{2}}=\mathbb{E}_{\epsilon\sim\mathcal{N}(0,1),\,t}\left[\alpha*\left\|\epsilon-\epsilon_{t}^{Albedo}\right\|_{2}^{2}\right]
\end{equation}
To alternately optimize $\mathcal{L}{_\text{1}}$ and $\mathcal{L}{_\text{2}}$, we set the balancing factor $\alpha = 2$ to smooth the loss curve during training.

%% file: section/5.Experiments.tex
\section{Experiments}
\label{sec:exp}

\subsection{Experimental Setting}
We select over \textbf{100K} PBR-textured meshes from Objaverse and \textbf{90K} from TexVerse as our general 3D training dataset. In addition, we use more than \textbf{150K} diffuse-textured meshes from Bedlam to build 3D Garment Videos for cross-pose augmentation. To enable cross-topology texture learning, we sample frames from the 3D Garment Videos and use the Nano Banana to generate about \textbf{50K} edited images.

Our diffusion backbone is based on Stable Diffusion 2.1, trained on 8 H200 GPUs for about 10 days with a batch size of 2 and an image resolution of 512×512. For evaluation, we use both industrial well-rendered images and wild images from DeepFashion2~\cite{ge2019deepfashion2} masked by SAM2~\cite{ravi2024sam} as input prompts, and select industrial 3D meshes together with Hunyuan3D-generated meshes as target geometries.

\noindent\textbf{Baselines.}
We compare our results under both geometries with several classic or state-of-the-art commercial models, including Paint3D, Hyper3D OmniCraft Texture Generator (Deemos Rodin), Hunyuan3D (Tencent), and Meshy 6 Preview (Meshy AI). Among them, Paint3D outputs only albedo maps, while the others generate both albedo and MR maps. Except for Paint3D, all model results are sourced from their official websites via paid access or granted credits.

\begin{figure}[t]
    \centering
    \includegraphics[width=\linewidth]{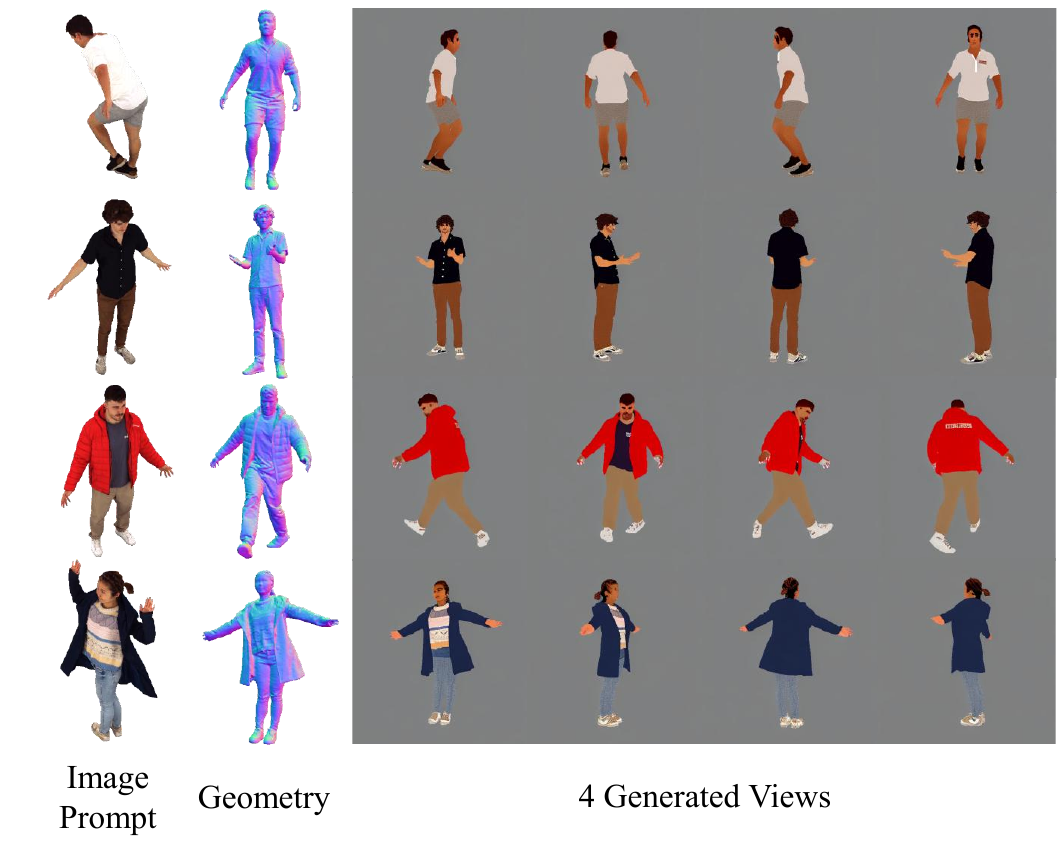}
    \caption{We use 4D-Dress as our test set to evaluate the model’s ability to generate textures across different human poses while maintaining consistent garment topology.}
    \label{fig:Cross-pose}
\end{figure}

\subsection{Qualitative Evaluation}
\label{sec:qualitativeEvaluation}

\begin{figure*}[t]
    \centering
    \includegraphics[width=\linewidth]{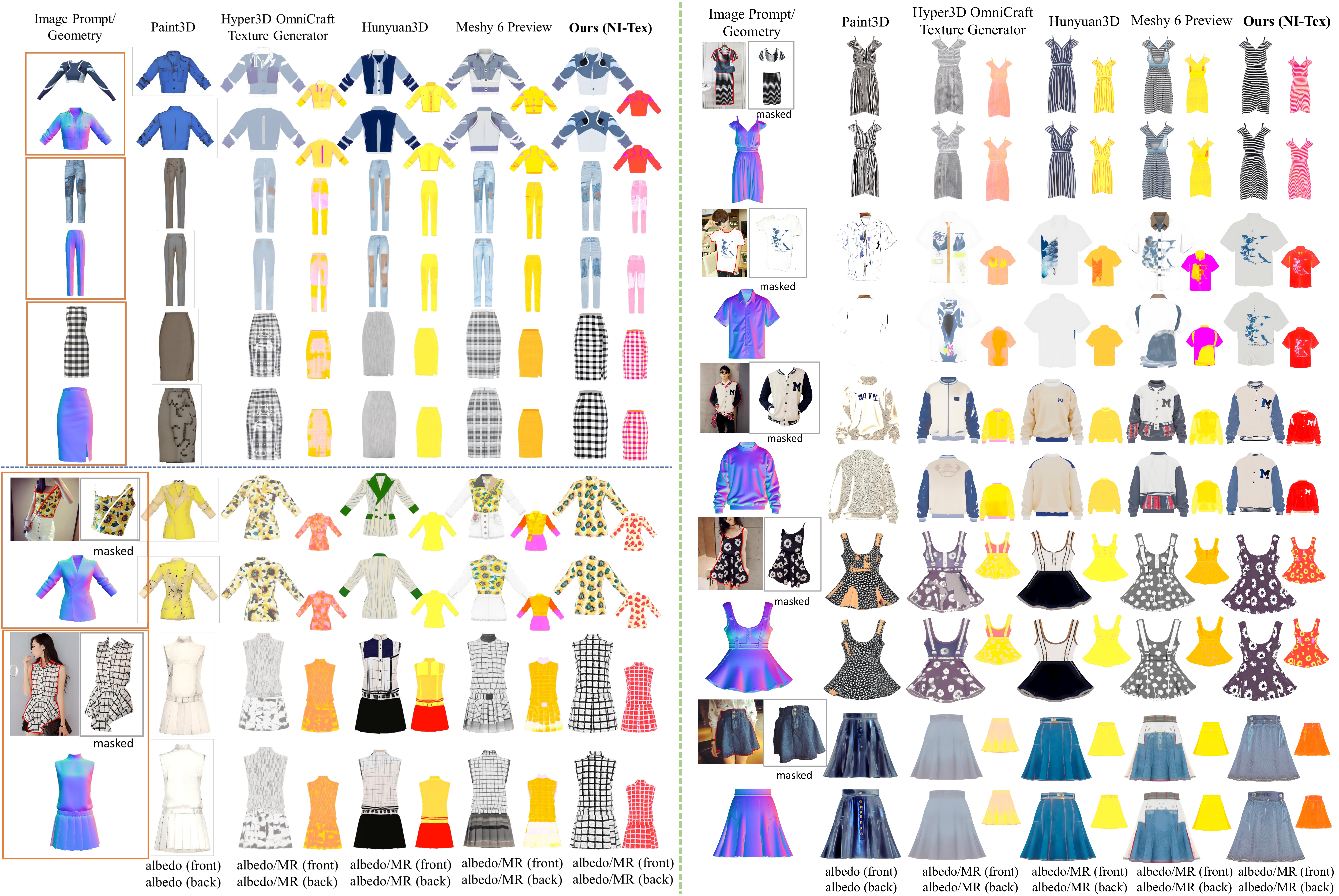}
    \caption{\textbf{(left)} NI-Tex results using industrial well-rendered images (top) and DeepFashion2 wild images masked by SAM2 (bottom) as prompts, with industrial meshes as targets. NI-Tex produces realistic industry-level PBR materials and aligns textures across complex garment surfaces. The top also shows robust cross-topology generation under consistent poses. 
    \textbf{(right)} NI-Tex results using DeepFashion2 wild images as prompts with generated meshes as targets, demonstrating reliable PBR material generation.
    }
    \label{fig:Industrial Mesh}
\end{figure*}

\noindent\textbf{Generation for Industrial Meshes.}
We perform experiments with two categories of input images in Figure~\ref{fig:Industrial Mesh} (left). The first group (top), named as well-rendered images, is rendered from existing 3D assets and are aligned with the target meshes in pose. This setting is specifically verify the effectiveness of the \textbf{cross-topology texture generation} in NI-Tex. The second group (bottom) of input image prompts is sourced from DeepFashion2, comprising real-world and web-collected images whose poses and geometries exhibit large discrepancies from the target 3D meshes. Despite this gap, NI-Tex successfully captures the correct texture information.

\noindent\textbf{Generation for Generated Meshes.}
We generate meshes using Hunyuan3D text prompts. Compared to industrial meshes, these generated meshes exhibit more folds, and their poses and geometries present significant gaps relative to the target images from DeepFashion2. Despite this, NI-Tex produces textures that not only closely align with simple sewing patterns, but also faithfully preserve structural garment details (see Figure~\ref{fig:Industrial Mesh}, right).

\noindent\textbf{Effectiveness of Cross-pose Texture Generation.}
The 4D-Dress dataset~\cite{wang20244d} provides per-frame 3D captures of real human motion, though its texture maps often contain baked shadows. We use 4D-Dress to maintain consistent garment topology across poses, enabling a focused evaluation of NI-Tex’s cross-pose texture generation (See Figure~\ref{fig:Cross-pose}). For each sequence, we randomly sample a mesh and viewpoint to render the input image prompt, and use another pose of the same subject as the target mesh.

\noindent\textbf{Baking Strategy.}
We compare our baking strategy with the coverage-based baking strategy on multiple complex garment meshes qualitatively in Figure~\ref{fig:DemoForViewSelect}. We quantitatively compare the two baking methods using metrics $i^{\textup{UQ}}$ and $i^{\textup{cvg}}$.
\begin{equation}
    i^{\textup{UQ}} = \arg \max_{i} {\sum_{p \in \mathcal{P}_i} U_p},~\quad i^{\textup{cvg}} = \arg \max_{i} {\sum_{p \in \mathcal{P}_i} V_p}
\end{equation}
where $\mathcal{P}_i$ is the set of pixels for view $i$, $U_p$ is the predicted uncertainty of pixel $p$ and $V_p=1$ only if $p$ is not covered by any views. Due to space limitations, quantitative results and more qualitative experiments are provided in the appendix.

\subsection{Quantitative Evaluation}
To quantitatively verify the effectiveness of NI-Tex in Table~\ref{tab:quantitative_results}, we adopt the Kernel Inception Distance (KID) metric, which provides an unbiased estimate similar to FID, and also report FID for reference. We randomly select 10 textured objects from both industrial and generated meshes, render them from multiple viewpoints, and perform KID experiments using a fixed set of 42 random seeds.

\begin{table}[h]
\caption{We compare Paint3D, Hyper3D OmniCraft Texture Generator, Hunyuan3D, Meshy, and NI-Tex using 10 image prompts and multiple rendered views of 10 textured objects in terms of KID and FID. NI-Tex achieves the best performance on both metrics.}
\label{tab:quantitative_results}
\centering
\resizebox{\linewidth}{!}{
\begin{tabular}{lccccc}
\toprule
Method & Paint3D & Hyper3D & Hunyuan3D & Meshy & Ours(NI-Tex) \\
\midrule
KID $\downarrow$ & 0.0695 & 0.0471 & 0.0528 & 0.0383 & \textbf{0.0364} \\
FID $\downarrow$ & 293.45 & 285.45 & 272.34 & 246.39 & \textbf{237.59} \\
\bottomrule
\end{tabular}
}
\end{table}

%% file: section/6.Conclusions.tex
\section{Conclusions}
\label{sec:conclusions}
In summary, we propose a novel technique, NI-Tex, for non-isometric image-based garment texture generation. By constructing 3D Garment Videos and applying Nano Banana to edit images from videos, we create image–garment training pairs that enable NI-Tex to perform stable cross-pose and cross-topology texture generation. To produce complete textures on the mesh, we employ iterative baking with view selection and uncertainty quantification. Extensive experiments demonstrate that the generated materials are suitable for both industrial graphics design and commercial 3D generation models.

\noindent\textbf{Limitations and Future work.} 
While NI-Tex delivers high-quality non-isometric texture generation, its generalization to complex rigid deformations remains limited due to the lack of physically simulated data for general objects. In future work, we aim to enhance the model’s 3D self-awareness of object deformation, enabling more robust non-isometric texture generation under limited data.

%% file: section/Appendix.tex
\clearpage
\setcounter{page}{1}

\section{Implementation Details}
\label{sec:AppendixImplementationDetails}


\subsection{MR Rectification for Cross-pose Supervision}
\begin{figure}[h]
    \centering
    \includegraphics[width=0.8\linewidth]{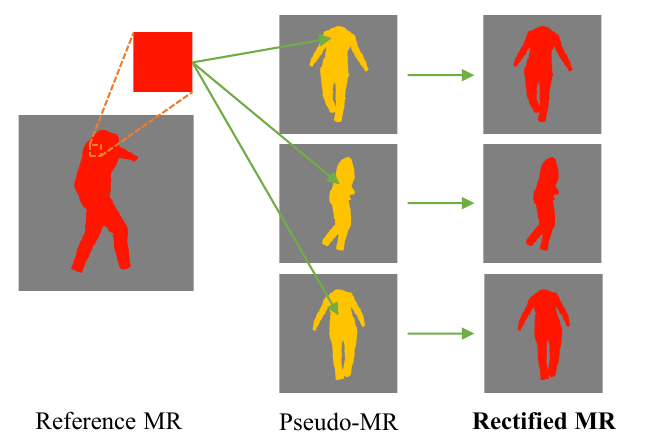}
    \caption{We randomly select one MR image from the condition 3D frame as the reference MR image. Since each MR map is assigned a globally uniform value, we can extract any foreground pixel as the reference pixel. We then index the MR images of the supervision 3D frame and replace all foreground pixels with the reference value, enabling cross-frame supervision.}
    \label{fig:MRRexDemo}
\end{figure}

Since the metallic and roughness values vary across frames, their reflective behaviors are inconsistent, making direct supervision from the supervision 3D frame unreliable. To address this, we introduce MR rectification (Figure~\ref{fig:MRRexDemo}). We sample a representative foreground pixel from the MR image of the condition 3D frame and replace all foreground regions of the supervision frame’s MR images with its value, enabling consistent cross-frame supervision during training.


\subsection{Iterative Baking Algorithm Details}
In this subsection, we describe the implementation details of the baking algorithm. Specifically, we talk about the training of UQ model, the details of the reweighting of multi-view texture maps.

\paragraph{UQ Model Training.} We adopt the resnet-50 architecture that predicts per-pixel uncertainty from an single-view rendered texture map.
The training dataset is constructed from pairs of rendered texture map (intermediate baking result from the baking algorithm) and the ground truth texture map of the same view. The data pairs come from an an error simulation procedure describe in section~\ref{subsec:baking}. To quantify the difference between the rendered texture map and the ground truth texture map, we use the SSIM metric to quantify the per-pixel uncertainty map from two texture maps. Specifically, the compute the per-pixel SSIM values are in the range $[0, 1]$. The supervision loss is simply:
\begin{equation}
    \sum_{p_i} \|\textup{UQ}(p_i) - y^{\textup{SSIM, GT}}(p_i)\|_2^2
\end{equation}
where $p_i$ is a pixel, $\textup{UQ}(p_i)$ denotes the predicted uncertainty value and $y^{\textup{SSIM, GT}}(p_i)$ is the SSIM value computed by comparing with the ground truth texture map.

\paragraph{Multi-view Reweighting.} The UQ model is also used to reweight texture maps based on the predicted uncertainty scores. When performing multi-view blending, we compute the final texture $t_i^{\star}$ from the texture map of multiple views $\{p_{ij} | j\}$, where each view $j$ is weighted by the uncertainty score $(1-\textup{UQ}(p_{ij}))$ and a constant view score $c_j$ corresponding to how far the view is from the frontal and back view. For the front and back viewpoints, we set $c_j = 1$. For other viewpoints, $c_j$ is progressively attenuated to $0.5$, $0.25$, $0.125$, and $0.1$ according to their relative distance and perceptual importance.

\begin{equation}
    t_i^{\star} = \frac{\sum_{j} (1-\textup{UQ}(p_{ij})) c_j p_{ij}}{\sum_{j} (1-\textup{UQ}(p_{ij}))  c_j + \epsilon_1}
\end{equation}

\subsection{Data Preparation Details}
We encode roughness and metallic into an RGB image, where the R channel is fixed to 255, and the G and B channels store the corresponding values scaled by 255. For each frame, we randomly apply different types of lighting sources, including point lights, area lights, and environment lights. We render the garment under 10 views, capturing the illumination effects together with material-related attributes (e.g., PBR texture properties) and geometry-related attributes (e.g., normal images and position images).

Since the MR values in our 3D garment videos are globally uniform, the model tends to overfit these constant properties during training. To enhance the network’s perception and generalization across different MR materials, we incorporate additional supervision from the Objaverse and TexVerse datasets for cross-mixed training.

\begin{figure*}[h]
    \centering
    \includegraphics[width=\linewidth]{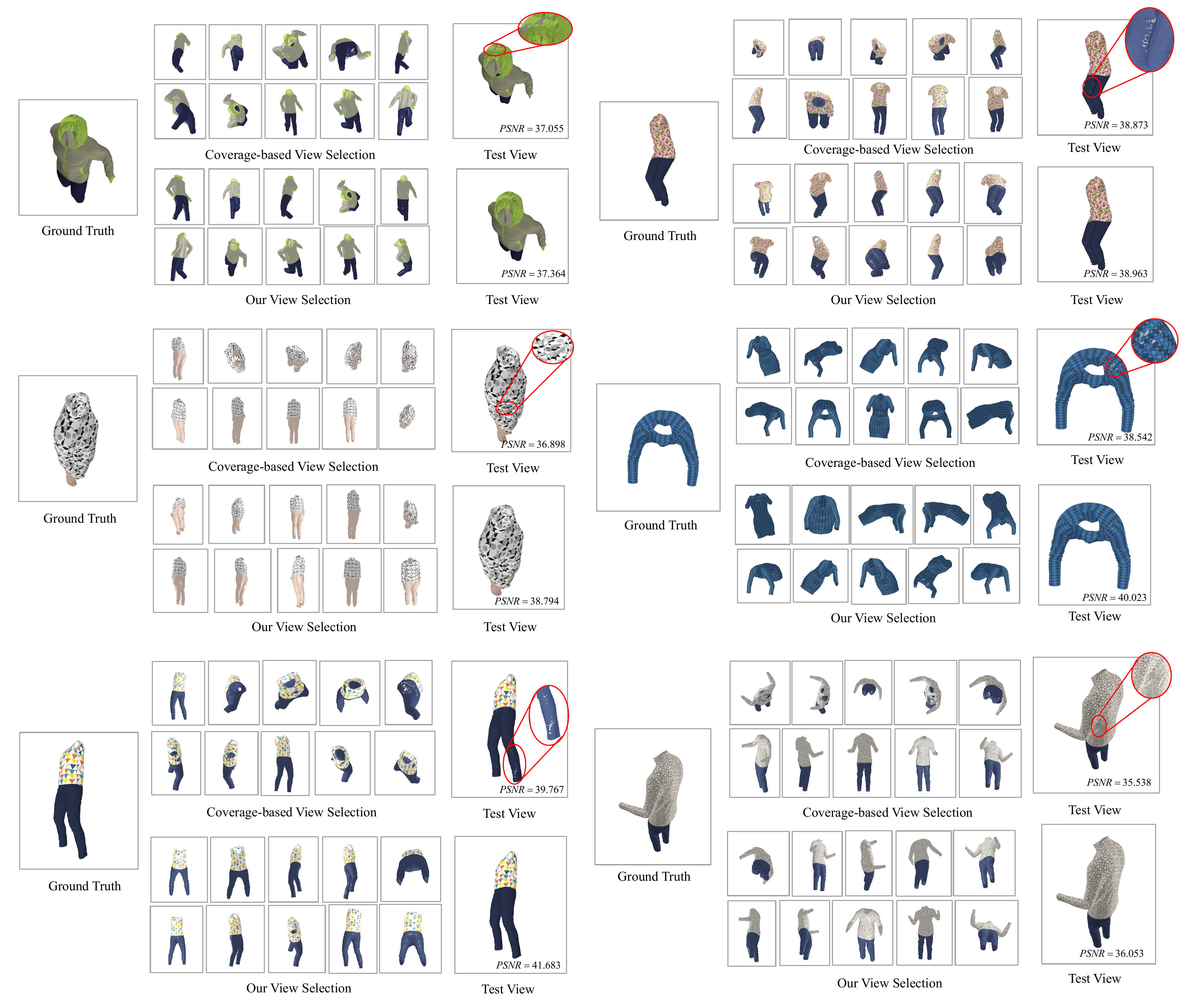}
    \caption{We compare the 10 views selected by the coverage-based strategy with those selected by our strategy. We then choose the worst viewpoint as the test view and compute its PSNR.}
    \label{fig:SupViewSelection}
\end{figure*}

\begin{figure*}[t]
    \centering
    \includegraphics[width=0.75\linewidth]{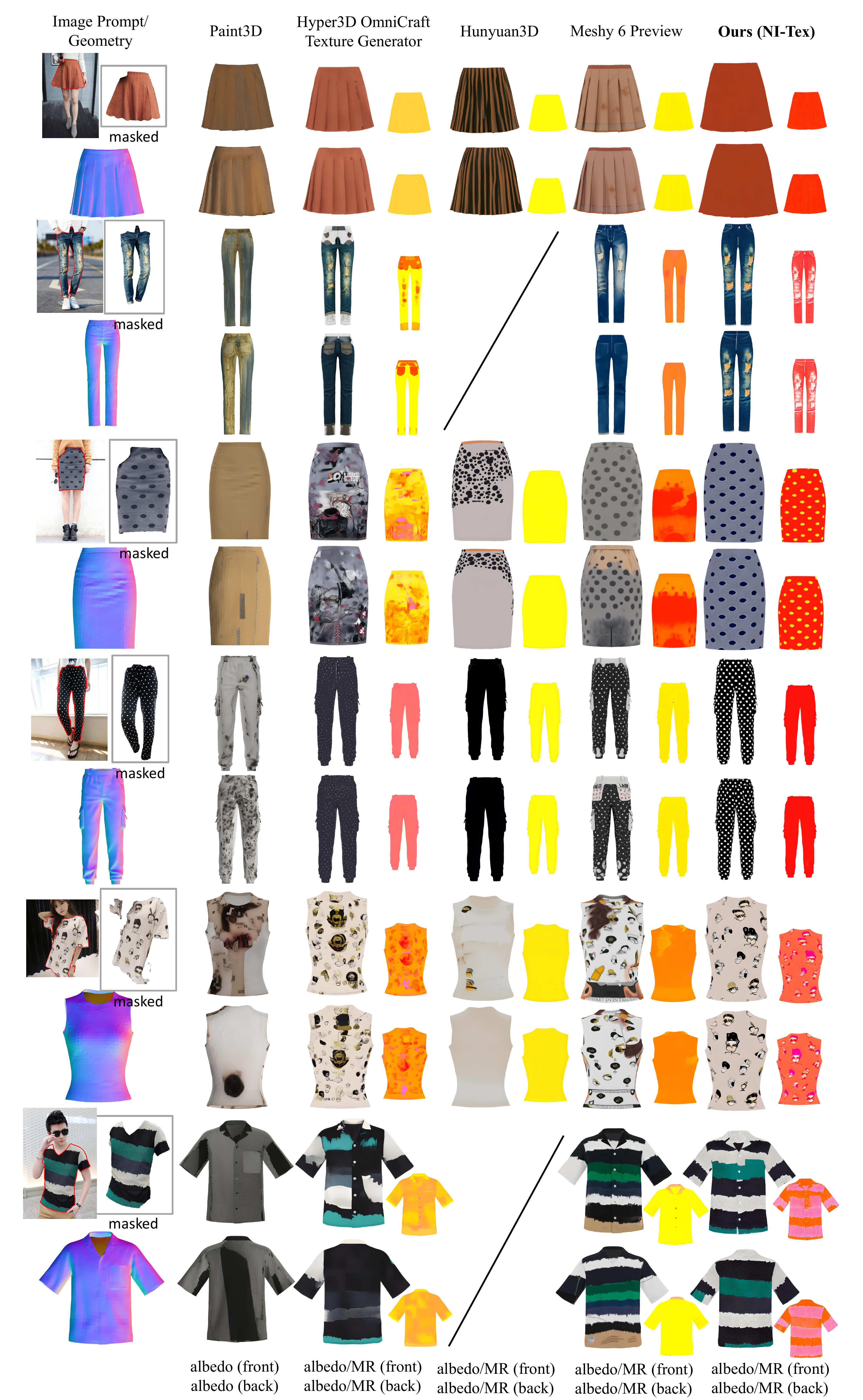}
    \caption{Texture generation results on industrial meshes using wild images from DeepFashion2 as image prompts. NI-Tex effectively captures the appearance of the input images, even under challenging variations. ('/' indicates generation failure.)}
    \label{fig:Industrial Mesh Sup}
\end{figure*}

\begin{figure*}[t]
    \centering
    \includegraphics[width=0.78\linewidth]{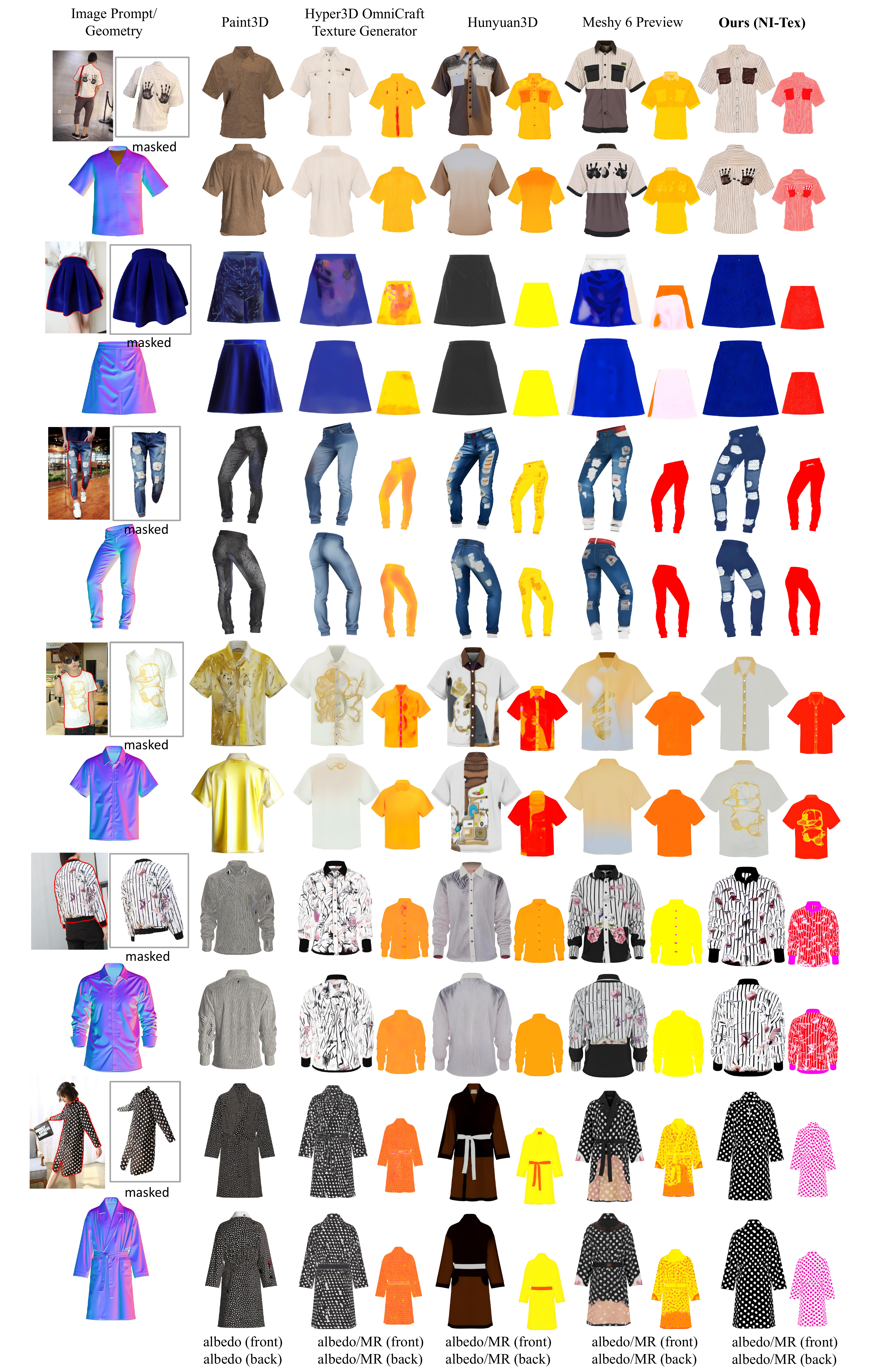}
    \caption{Texture generation results on Hunyuan-generated meshes using wild images from DeepFashion2 as image prompts. NI-Tex demonstrates strong capability in retaining fine-grained details, including logos and intricate patterns, while accurately capturing textures across diverse clothing types.}
    \label{fig:Generated Mesh Sup}
\end{figure*}

\begin{figure*}[t]
    \centering
    \includegraphics[width=0.82\linewidth]{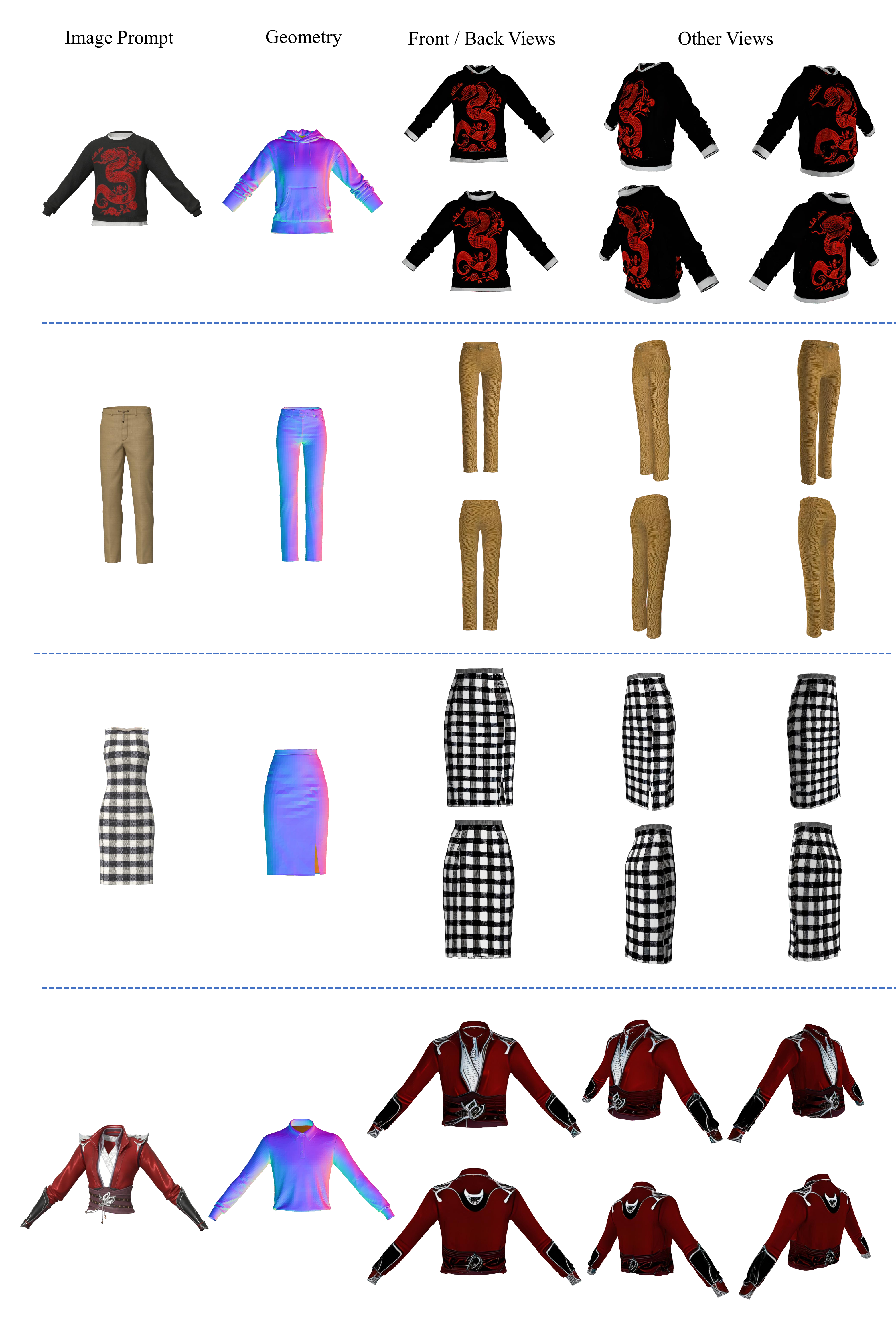}
    \caption{Multi-view visualization for industrial meshes (using well-render image prompts).}
    \label{fig:SupMultiviewResults1}
\end{figure*}

\begin{figure*}[t]
    \centering
    \includegraphics[width=0.78\linewidth]{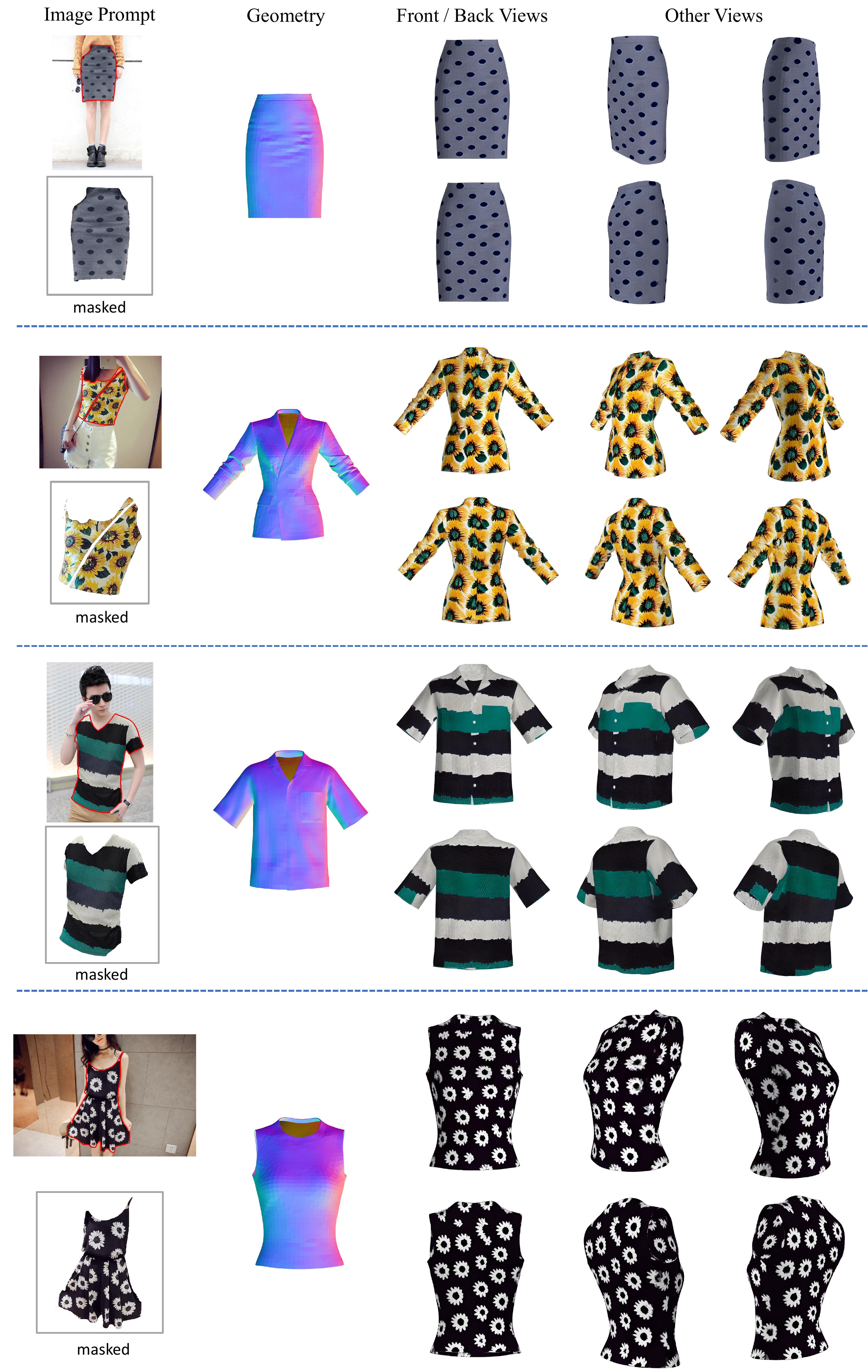}
    \caption{Multi-view visualization for industrial meshes (using image prompts from DeepFashion2).}
    \label{fig:SupMultiviewResults2}
\end{figure*}

\begin{figure*}[t]
    \centering
    \includegraphics[width=0.76\linewidth]{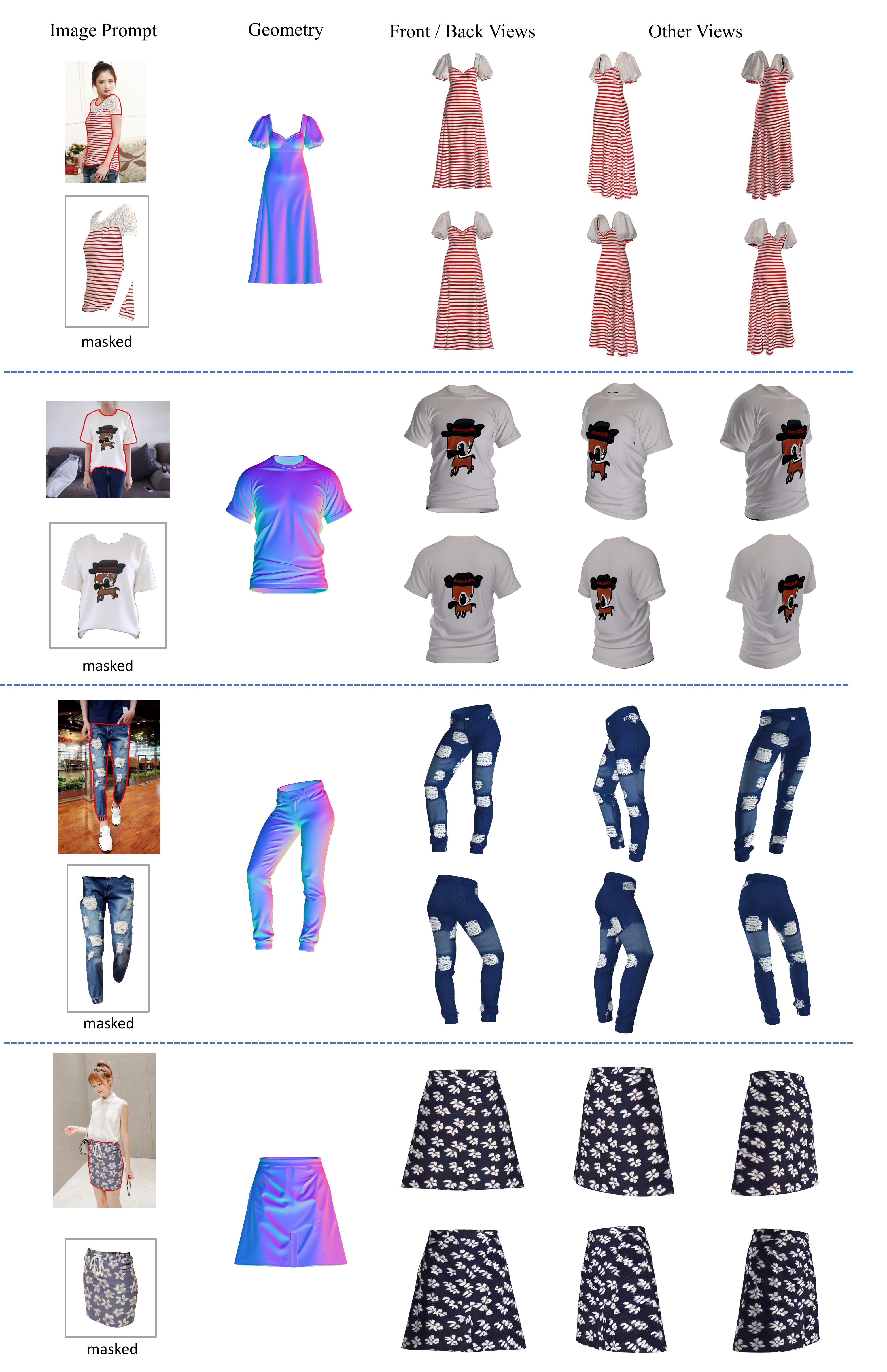}
    \caption{Multi-view visualization for Hunyuan-generated meshes (using image prompts from DeepFashion2).}
    \label{fig:SupMultiviewResults3}
\end{figure*}

\section{Experiments}
\subsection{Generation for Industrial Meshes}
In Figure~\ref{fig:Industrial Mesh Sup}, we supplement additional examples using wild images from DeepFashion2 as image prompts, with industrial meshes as the target meshes. We find that NI-Tex is capable of reliably generating textures that closely conform to the input image prompts, maintaining high fidelity even in highly challenging conditions.

\subsection{Generation for Generated Meshes}
In Figure~\ref{fig:Generated Mesh Sup}, we present additional examples using wild images from DeepFashion2 as image prompts, with Hunyuan-generated meshes as the target geometry. We observe that NI-Tex effectively preserves logos and local details, and can also faithfully maintain complex patterns such as spots.

\subsection{Multi-view Visualization}
To further verify the 3D consistency of the textures generated by NI-Tex, we project the texture maps back onto the mesh surface. In addition to the front and back views, we further show four additional viewpoints: front-left, back-left, front-right, and back-right. Results in this subsection are displayed under lighting conditions and some of the examples are taken from previous experiments. (See Figure\ref{fig:SupMultiviewResults1}, \ref{fig:SupMultiviewResults2}, \ref{fig:SupMultiviewResults3}
)

\subsection{Baking Strategy}
After training the UQ model, we compare our view selection strategy with the coverage-based view selection baseline. We separately use the two greedy metrics, $i^{\textup{UQ}}$ and $i^{\textup{cvg}}$, to select 10 views on the given Bedlam geometry and bake. We select the worst viewpoints as our test views and perform quantitative evaluation using PSNR. Figure~\ref{fig:SupViewSelection} shows the 10 selected views for each strategy, along with their corresponding final test views and PSNR values.